# Mean Shift Rejection: Training Deep Neural Networks Without Minibatch Statistics or Normalization

Brendan Ruff and Taylor Beck and Joscha Bach[1]

**Abstract.** Deep convolutional neural networks are known to be unstable during training at high learning rate unless normalization techniques are employed. Normalizing weights or activations allows the use of higher learning rates, resulting in faster convergence and higher test accuracy. Batch normalization requires minibatch statistics that approximate the dataset statistics but this incurs additional compute and memory costs and causes a communication bottleneck for distributed training. Weight normalization and initialization-only schemes do not achieve comparable test accuracy.

We introduce a new understanding of the cause of training instability and provide a technique that is independent of normalization and minibatch statistics. Our approach treats training instability as a *spatial common mode signal* which is suppressed by placing the model on a *channel-wise* zero-mean isocline that is maintained throughout training. Firstly, we apply channel-wise zero-mean initialization of filter kernels with overall unity kernel magnitude. At each training step we modify the gradients of spatial kernels so that their weighted channel-wise mean is subtracted in order to maintain the common mode rejection condition. This prevents the onset of mean shift.

This new technique allows direct training of the test graph so that training and test models are identical. We also demonstrate that injecting random noise throughout the network during training improves generalization. This is based on the idea that, as a side effect, batch normalization performs deep data augmentation by injecting minibatch noise due to the weakness of the dataset approximation.

To compare with many previously published results, we demonstrate our method with the standard ResNet110 and CIFAR-10 image classification task and training protocols. Our technique achieves higher accuracy compared to batch normalization and for the first time shows that minibatches and normalization are unnecessary for state-of-the-art training. We believe the method is applicable generally to the training of any deep neural network including recurrent designs. We further show a fatal problem with L2 weight decay and replace it with an equivalently weighted unity magnitude decay that controls both filter inflation and deflation.

**Preface to arxiv edition.** This pre-print of the conference submission is revised with reference to the recent arxiv publication of *Filter Response Normalization* which is an alternative method that also avoids minibatch statistics to improve upon the performance of batch normalization. Due to the high relevance, we compare and contrast our work to theirs and have started additional experiments using their example architectures and tasks.

## 1 INTRODUCTION

It is generally accepted that a high learning rate is required for improved generalization and best test accuracy, and this requires normalization of one form or another to achieve state of the art accuracy. Attempts to avoid normalization using novel weight initialization schemes [25] have failed to deliver equivalent test accuracy and have required *ad hoc* architecture changes. Normalization, and in particular the use of minibatch statistics, has many drawbacks including increased computational cost, larger memory footprint, and communication overhead for distributed training over multiple processors. Despite these drawbacks, batch normalization has yet to be replaced with an equally effective alternative until very recently [26] and so up to now has been the default technique for training deep neural networks. The root cause of training instability is far from proven or even understood, raising the question of whether another approach could succeed while avoiding normalization and minibatch statistics entirely.

Recently, *Filter Response Normalization* [26] has avoided the use of minibatch statistics and presented higher accuracy results than the batch normalization baseline. This requires activation magnitude normalization within a single sample and so does not avoid normalization. The results are impressive and we believe match our own though we cannot directly compare without further work.

It has been shown that the mean shift component of covariate shift ([4], [12], [18]) may be controlled with mean subtraction using minibatch statistics. Without such control training is adversely affected. But can it be eliminated at its root?

We re-examine the cause of training instability and in particular mean shift, taking inspiration from *common mode signal rejection* in networks of analog electronic amplifiers. During training, a DNN behaves very much like an analog network in which the operational amplifiers are analogous to convolutional filters. They experience common mode noise at their inputs, just like mean shift. If left unchecked, this causes instability in the forward signal as the mean signal is amplified, just like CNNs. In particular, analog amplifiers use a differential signal at their input that removes the common mode signal which otherwise would be highly amplified. This is our main inspiration and insight. We believe and will demonstrate that the training instability in DNNs is due to the spatial common mode signal alone, and therefore this can be cured without resorting to minibatch statistics or normalization.

[1] AI Foundation, San Francisco, USA
email: brendan|taylor|joscha@aifoundation.com

A spatial filter has x-y extent (e.g. Dx3x3 where 3x3 is the spatial extent and D the depth), whereas a pointwise filter that is 1x1 has no spatial extent other than the single point. The input from a previous $l\text{-}1$ layer's $k^{th}$ filter may be thought of as a base signal $_{l-1}x_k$ that is spatially varying to which is added a constant mean shift $_{l-1}S_k$ that is not spatially varying, at least locally within the spatial extent of the filter kernel. $_{l-1}S_k$ is shared spatially across the weights $_lW_{f,k,i}$ which is the $k^{th}$ 2D slice of filter $_lW_f$ whose weights within a 2D slice are indexed spatially by $i$. $_{l-1}S_k$ is the channel-wise spatial common mode signal in the $k^{th}$ input to layer $l$, and this is shared across all filters in the layer. Batch normalization counters this shift in the previous layer by subtracting the minibatch mean, and so in this sense $_{l-1}S_k$ corresponds to the minibatch mean.

If the $k^{th}$ filter in the previous layer $l\text{-}1$ exhibits a mean shift $_{l-1}S_k$ in its output $_{l-1}x_k^{shifted}$ that disturbs it from the unshifted output $_{l-1}x_k$ so that

$$_{l-1}x_k^{shifted} = {_{l-1}x_k} + {_{l-1}S_k}$$

then the next layer's filter indexed by $f$ will manifest a shift in its output $_lx_f^{shifted}$ due to the input shift so that

$$_lx_f^{shifted} = {_lx_f} + {_lW_{f,k,i}} * {_{l-1}S_k}$$

where $_lW_{f,k,i}$ is the $k^{th}$ 2D slice of the $f^{th}$ filter kernel $_lW_f$ in layer $l$ and $i$ indexes spatially within the 2D slice, and $*$ is the convolution operator over $i$. We note that if the 2D slice $_lW_{f,k}$ is arranged to have zero mean over $i$, then the convolution over $i$ is undisturbed since $_{l-1}S_k$ is a constant term for at all positions $i$ and $\sum_i {_lW_{f,k,i}} = 0$. Hence a filter that has channel-wise zero mean (CZM) is immune to mean shift in its inputs. Any mean shift from a previous layer has no impact and therefore mean subtraction is not necessary in that previous layer. This also applies to a previous pointwise layer or any other input as each filter performs *mean shift rejection* no matter the source of the input. Note that we do not apply this to the first layer in the net that is connected to the input e.g. RGB image.

If the channel-wise zero mean (CZM) condition is slightly relaxed, the filter is still highly resistant to mean shift in its inputs. The question is whether a DNN initialized with CZM will remain in the approximately CZM condition throughout training. In practice we find that it does, but there needs to be some encouragement for CZM to remain sufficiently stable for the optimal result.

The CZM condition defines an isocline within the model space. Any departure from this isocline may result susceptibility to mean shift, which naturally propagates through the net and increases the final loss. We discover that the CZM state is inherently stable during training and we make the observation that there is negative feedback to maintain the CZM state due to the local loss minimum in which it sits. We currently lack rigorous analysis of this phenomenon and leave it to future work. However, we believe the logical argument is compelling and the empirical results support it.

We suspect that the negative feedback introduces oscillation around this local minimum, which we believe is detrimental to optimal training. To reduce this oscillation, we introduce a modification to the gradients in the weight update step outside of the computational graph. As described in section 3.3, we subtract the channel-wise mean of the gradients from the gradients of the weights of each 2D kernel slice which we term channel-wise zero mean gradients (CZMG). By applying this CZMG at each training step, the model remains on the CZM isocline without oscillation. Since spatial convolutional layers in the CZM state are immune to mean shift, and as pointwise layers are typically interspersed with spatial layers, then any mean shift in the pointwise layers is rejected by the next spatial layer and thus the entire model is immune to mean shift.

We must ask whether the model can train effectively when limited to move within this CZM isocline. We find that in practice it can, but in some cases we need to relax the CZM condition. We add a *zmg* factor to the CZMG subtraction (e.g. 0.85). Though this is a hyper parameter, we have found 0.85 to 0.9 works for a variety of nets that we have tried.

Despite the plethora of weight initialization methods, we find that initialization of filters by a random uniform distribution of unit magnitude, adjusted after the CZM subtraction is applied, as effective as *fanin* adjusted methods such as He's [10] and Glorot's [8] and Fixup [25].

As with weight normalization, we reparametrize the filter kernel with an additional trainable scaling factor that is exponentiated. We note that without exponentiation the scaling factor can become negative, so the parameter must train quickly to cross through zero from its initial value of 1. We attribute this to the random filter kernel initialization being randomly inverted for some filters.

During training, filters tend to inflate in magnitude [14] and this is commonly countered by applying L2 weight decay. However, the interplay between the scaling and kernel magnitude in the presence of weight decay causes a dynamic effective learning rate even without normalization due to the relative learning rate step size compared to the filter magnitude. We note that this frequently leads to a runaway deflation of filters whose effective learning rate therefore experiences a spiraling increase. This leads to filters that can no longer learn effectively as their update step moves the model too far in each training step as the weight decay drives their magnitude towards zero. With CIFAR-10 and ResNet110, the baseline batch normalization training typically loses around 25% of its filters to this condition so reducing the useful capacity of the net.

To counter this problem, we replace the L2 weight decay with *L2 unity magnitude anchoring* (LUMA). The intuition is that the magnitude of the filters in each layer throughout the entire network should remain near their unity initial value during training and only their direction should change. As with weight normalization, scale is controlled by the separate parameter which has no decay. Unity anchoring also acts as a lateral inhibition between channels in a filter, so if one weight needs to increase in magnitude then less important weights must reduce. This naturally leads to sparser connectivity which is known to reduce over-fitting and improve generalization [13]. We assign LUMA the same loss weighting as the replaced L2 decay so there are no additional hyper parameters to search. An analysis of the final model shows that all filters remain very close to unity magnitude.

Finally, we speculate that the effectiveness of minibatch statistics in training generalization is in part due to the noise it introduces into the forward signal since the mean and variance of each randomly chosen minibatch is different. Ad hoc we introduce a multiplicative modulation to the forward signal with random uniform distribution with unit mean and 0.1 amplitude. This gives significant improvement in final accuracy.

To summarize our contributions:
- We introduce channel-wise zero mean initialization (CZMI) for spatial filters that places the model onto a CZM isocline that allows high learning rate training by making the filters immune to mean shift.

- We introduce weighted channel-wise mean gradient subtraction that encourages the model to remain close to the channel-wise zero mean isocline throughout training.
- We demonstrate the effectiveness of a unity magnitude uniform random distribution in the filter initialization.
- We replace L2 weight decay with an L2 unity magnitude anchor that is loss weighted as with L2 weight decay.
- We demonstrate direct training of the test graph at high learning rates that outperforms batch normalized training.
- We demonstrate that deep noise injection benefits generalization which we note is an unintentional side effect of batch normalization.

## 2 RELATED WORK

The two most popular and effective techniques for state-of-the-art training of DNN's in common use are batch normalization [12] and weight normalization [18]. Batch normalization relies on minibatch statistics as an approximation to the dataset statistics. Weight normalization performs best if minibatch mean subtraction is applied. Weight normalization reparametrizes the convolutional filter kernel to separate the magnitude of the kernel from its direction but does not achieve the same performance as batch normalization. We adopt their exponentiated scaling variant mainly to avoid negative scaling values. In addition, it naturally prevents the scaling from overly reducing. The reason is the asymmetric size of a positive step compared to a negative step for the exponentiated scale given a linear training model such as backpropagation.

The use of minibatch statistics allows for best final accuracy and model generalization but at the cost of increased computation and memory footprint [5] and is incompatible with many network architectures (e.g. [6] and [17]) and methods of training [23]. Additionally, when normalization is combined with first order training such as backpropagation and in particular when weight decay is applied, the local effective learning rate of filters varies enormously [14] and so this dynamic and varying local effective learning rate becomes entangled with the choice of the initial learning rate and the annealing strategy. In contrast, we neither apply normalization nor weight decay and so do not encounter the effective learning rate issue.

Without normalization it is found that the learning rate must be reduced to avoid gradient explosion but this does not achieve comparable test accuracy and is slower to train. Attempts to support higher learning rates by using more stable model initialization schemes like Fixup [25] do not achieve comparable accuracy and require changes to the network architecture. Notably, Fixup trains with a learning rate lr=0.1 but the effective learning rate with batch normalization is 0.4 which is 4x higher. In contrast, we train our method with a learning rate of 0.4, as in our method the effective and actual learning rate are the same and do not vary in training.

Very recently, *Filter Response Normalization* [26] has avoided the use of minibatch statistics and presented higher accuracy results than the batch normalization baseline. However, this still requires normalization of the activation magnitude within a single sample. This spatially couples the response within an activation map which imposes a lateral inhibition regularizer on the training. It is far from clear whether that is a good thing and what other artefacts that could impose on the model, though clearly the method achieves better results than batch normalization on their extensive baselines. However, they omit the simpler CIFAR task, and without that we cannot compare directly. Quite likely they would achieve similar results to our method as we believe it is the mean subtraction that is the main cause for concern with respect to minibatch statistics, and this omission is common to both methods. To address this comparison issue, we are actively engaged in running baselines on some of their results. It may be that their magnitude normalization has a very similar regularizing affect compared to our unity magnitude anchoring though ours is both less computationally expensive and does not require statistics to be collected across the activation map which could be problematical with large input maps.

Also, very recently [27] has shown that eliminating singularities in training with minibatch statistics allows for a micro minibatch to be employed with improved results upon a batch normalization baseline. It would be interesting to analyze our method with this.

For the small minibatch regime, Weight Standardization [28] demonstrates improvements over batch normalization but fails to beat with larger minibatches.

Regarding zero mean weight initialization, [29] demonstrated the application of zero mean weight normalization during training. However, they apply this across both spatial and channel dimensions whereas we consider the common mode signal to be only spatial and indeed applying it across the channel dimension is not effective. Also, we consider this to be a gradient *modifier* rather than a normalizer within the computational graph, though we concede that an alternative to the gradient normalization is to apply this channel-wise to the weights and so within the computational graph. As we found no benefit to this, then we chose the gradient modification approach as it lies outside of the computational graph and so cannot have regularizing side effects to the gradient computation.

Intense study into the training of deep convolutional neural networks has taken a particular focus on residual networks [10] as they demonstrate state-of-the-art performance and are applicable to a broad range of tasks. Much of the literature is devoted to the study of training ResNets both at initialization ([1], [8], [10], [9], [22], [20]) and during training ([24], [14], [18], [21], [7], [2], [15]). Batch normalization [12] still remains the leading training augmentation method for improving the generalization of deep convolutional neural networks ([11], [14], [21], [7]).

Though many works have proposed alternative theories ([4], [19], [11], [21]) for why batch normalization is so effective, the main explanation remains that a high learning rate explores the solution space more thoroughly, while minibatch statistics act as a regularizer that augments the dataset, which in combination improves generalization. We retain minibatch gradients where appropriate which is orthogonal to our method.

In particular, weight normalization [17], layer normalization [3], normalization propagation [2] and group normalization [16] are different approaches that focus on single sample statistics to avoid the computational and memory overhead of using minibatch statistics and to allow training with recurrent neural networks or online learning [6]. All non minibatch methods including weight normalization are noted to have worse performance than batch normalization [7] in comparable settings apart from the recent [26].

Other techniques build on top of batch normalization, such as generalized batch normalization [24] and batch renormalization [16] but do no match the baseline performance.

A well-known feature of batch normalization is its tolerance to large variation in choice of learning rate, which is thoroughly explored by van Laarhoven [14]. Our method is equally tolerant to choice of learning rate as we demonstrate whereas [26] requires a learning rate warm-up strategy.

Different to all previous techniques, we identify training instability as arising from a *common mode signal* that if left unchecked grows and prevents effective learning, and in particular the mean shift is identified as the culprit. Previous methods compensate for mean shift by applying mean subtraction or add additional bias parameters [25] [26]. Instead we initialize the model so that all filters lie on the channel-wise zero mean isocline and further maintain this condition throughout training by modifying the gradient update step by applying a weighted channel-wise zero mean subtraction. This achieves higher final accuracy than all previously published results without resorting to minibatch statistics or normalization, and moreover we directly train the test graph albeit with a reparameterization of the filter kernel using the exponentiated scaling variant as described in weight normalization [18].

## 3 TECHNICAL DESCRIPTION

### 3.1 Reparameterization

We reparametrize each convolutional filter kernel W (omitting subscripts for layer and filter index for clarity) according to

$$W = e^g \cdot V$$

where $g$ is a scalar and $V$ is the weights tensor. Note that the scaling parameter $g$ is exponentiated which prevents it from becoming negative and also reduces the gradient as the scaling reduces in size. This has a strong regularizing effect during training to prevent rapid reduction in its size and so has the effect of stabilizing the magnitude of the filter kernel *V*.

We find that $e^g$ must be initialized to be less than 1 for stability in early training. Typically, we set it to 0.8. Smaller values lead to faster and more consistent early training, while setting to 1 may cause the net to diverge after a few hundred steps.

Each convolutional neuron performs the computation

$$y = \Phi(W \cdot x + b)$$

where Φ is the nonlinear activation function (e.g. a linear rectifier) and b is the bias term if used. For inference of a trained model, the exponentiated scaling is combined into the weights tensor.

### 3.2 Channel-wise Zero Mean Initialization

We initialize each filter that has spatial dimensions greater than 1 as follows. First, we draw a uniform random sample X of the size of the weights tensor V.

$$X = U(-1, +1)$$

Next, we subtract from X the mean over the spatial dimensions, which for a typical 3D weights tensor is the two trailing x-y dimensions.

$$X' = X - E(X: dim = 1,2)$$

Finally, we normalize the Euclidean magnitude which retains the channel-wise zero mean state.

$$V = \frac{X'}{\|X'\|}$$

### 3.3 Channel-wise Zero Mean Gradients

We adapt the gradients ∇V for the weights tensor V by subtracting the weighted mean of those gradients over the spatial dimensions which (assuming a 3D weights tensor with two trailing spatial dimensions) is

$$\nabla V' = \nabla V - E(\nabla V: dim = 1,2) \cdot z$$

where $z$ is a constant hyper parameter. Typically, z=0.85 to 0.9 for ResNet110 and is not particularly sensitive to value. Note that this gradient update step is outside of the computation graph for the model. We call $z$ the *zmg* factor.

We find that with VGG like networks we can assign z=0.98 or even 1 with no difference in training rate or final accuracy, though with ResNets the *zmg* factor has a larger effect on final accuracy and training rate. In practice this is a constant set to 0.85 which reduces by 6.6x the gradient component normal to the CZM isocline.

### 3.4 L2 Unity Magnitude Anchoring

Rather than decaying all parameters including weights and biases and scaling to zero as with weight decay, we only decay the Euclidean magnitude of the filter kernels and set the decay target to unity rather than zero. For each filter this gives us an L2 unity magnitude loss $L_{LUMA}(V)$

$$L_{LUMA}(V) = \left(\sqrt{\sum_i V_i^2} - 1\right)^2 \cdot \lambda$$

where the subscripted $V_i$ is the i$^{th}$ weight of the weights tensor *V* and λ is the loss weighting e.g. 5e-4. We choose the loss weighting to be the same as the equivalent L2 weight decay.

## 4 EXPERIMENTS

To evaluate our approach, we choose ResNet110 and the CIFAR-10 dataset and follow the protocol of the original ResNet paper [10]. This DNN architecture is challenging due to its depth, and many studies ([19], [4], [11], [14], [18]) focus on this architecture and task and protocol to explore the operation of batch normalization and offer explanations for its efficacy. Our choice allows for a direct comparison of our results with the extensive research published on this topic and is quickly trained for exploration of hyper parameters.

The experimental protocol comprises training the net using backpropagation with stochastic gradient descent for 200 epochs at an initial learning rate of 0.1 and annealing by 0.1 at 100 and 150 epochs. The momentum is set to 0.9. Random scaling and flip augmentation is used as with He [10]. Using a minibatch size of 128, we execute 391 training steps per epoch and a total of 78200 steps. We repeat each experiment 8 times and record the mean and standard deviation. For the variations of our method we apply LUMA loss weighting to 5e-4 and for batch normalization we apply L2 weight decay of 5e-4.

### 4.1 Batch Normalization Baselines

We provide batch normalization baselines with L2 weight decay set to $5e^{-4}$ and a single learning rate of 0.1, which in all published studies achieves the highest final accuracy. Our results are comparable to previously published figures.

### 4.2 CZMI Baselines

We present channel-wise zero mean initialization only experiments (figure 1) without the gradient adaptation. We choose learning rates of 0.1, 0.2 and 0.4 to span the range from the batch normalization baseline of 0.1 to the estimated effective learning rate of 0.4.

### 4.3 CZMG Baselines

We combine CZMI and CZMG (i.e. CZMIG) and provide baselines (Figure 4) across a range of *zmg* factors of 0, 0.85, 0.98, and 1.0 with a learning rate of 0.4 which is optimal. This actual learning rate with filter magnitudes near unity is equivalent to the effective learning rate of 0.4 with batch normalization at actual learning rate of 0.1.

With batch normalization, we determine the average filter magnitude by inspecting the model parameters at the end of each epoch. After around epoch 10 the magnitude remains constant in the range 0.4 to 0.6 (typically with some exceptions in the earliest and deepest layers close to the classifier). Given filter magnitudes with a mean of around 0.5, then according to the inverse square law of the effective learning rate [22] this gives a mean effective learning rate of $0.1/0.5^2 = 0.4$. Hence our choice of 0.4 for training CZMIG.

We observe that the best *zmg* factor is 0.85 for this task and architecture. A *zmg* factor of 0.98 was found to be the highest that trains at a learning rate of 0.4. *zmg* factor of 1 was included for completeness with a learning rate of 0.1. This lower rate is shown because a higher learning rate diverges in this case.

### 4.4 Noise Injection Baselines

We include experiments with and without uniformly distributed noise injection of 0.1 amplitude and unit mean. This is applied at the input to the residual units. Hyper parameter search was used to determine the amplitude and we found this injection point to be the most effective for this architecture. We provide noise injection results at the target learning rate of 0.4 for direct comparison to the baseline batch normalization results.

## 5 RESULTS

**Table 1.** Summary of test accuracy in descending order across a range of training configurations.

| Method | LR | ZMG | Noise | Test Accuracy |
|---|---|---|---|---|
| **CZMIG** | 0.4 | 0.85 | 0.1 | **94.03 (±0.14)** |
| BatchNorm | 0.1 | 0 | 0 | 93.70 (±0.55) |
| CZMIG | 0.4 | 0.98 | 0.1 | 93.64 (±0.11) |
| CZMIG | 0.4 | 0.85 | 0 | 93.57 (±0.12) |
| CZMIG | 0.2 | 0.85 | 0 | 93.44 (±0.18) |
| CZMI | 0.4 | 0 | 0.1 | 93.33 (±0.11) |
| CZMI | 0.1 | 0 | 0 | 93.06 (±0.22) |
| CZMI | 0.2 | 0 | 0 | 92.95 (±0.23) |
| CZMI | 0.4 | 0 | 0 | 92.13 (±0.26) |
| CZMIG | 0.1 | 0.85 | 0 | 92.95 (±0.20) |
| CZMI | 0.1 | 1 | 0.1 | 91.03 (±0.25) |

The table above shows the accuracy of the various baselines in descending order for comparison. For CIFAR-10 and resnet110, CZMIG with lr=0.4 and *zmg*=0.85 with noise injection of 0.1 amplitude achieves the highest accuracy beating the batch normalization baseline by 0.33% which is significant. It is noteworthy that the error range for the model is far less with CZMIG and shows that the method is far more repeatable than batch normalization.

Interestingly, at the top of the error range it is shown that the best single batch normalization result matches that of the best CZMIG. Clearly a feature of batch normalization training is not that it cannot achieve a high accuracy, it is just that such high accuracy is not stable and so many experiments need to be run to get a single optimal model.

**Figure 1.** The training (top) and test (bottom) accuracies as a function of training progress, with channel-wise zero mean weight initialization

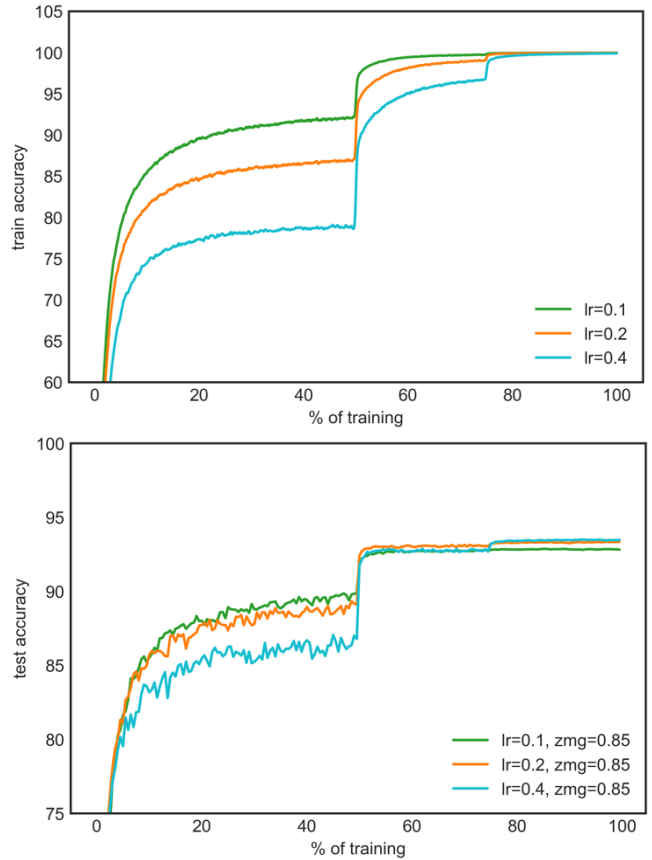

(CZMI) demonstrating the effect of different learning rate

Figure 1 shows that zero channel-wise zero mean weight initialization alone supports training of the ResNet110 at learning rates at and 4 times in excess of the base learning rate of 0.1 used with batch normalization. The best test accuracy is observed at a learning rate of 0.4 (93.33 ±0.11) with a considerable drop in accuracy at 0.1 (93.06 ±0.22). This is consistent with the idea that higher learning rate improves generalization. Also, the best result is considerably lower than the batch normalization and CZMIG baselines (table 1). However, channel-wise initialization alone is shown to be stable at all learning rates tried. Clearly, initialization alone is not sufficient for optimal training, a result shared by Fixup [25].

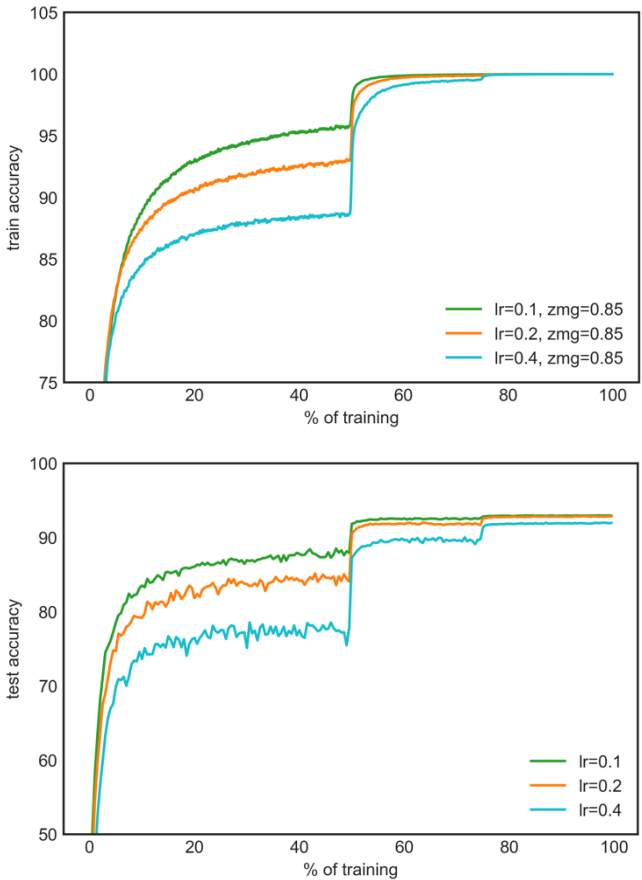

**Figure 2.** The training (top) and test (bottom) accuracies using CZMI combined with channel-wise zero mean gradient update (CZMG) demonstrating the effect of different learning rate

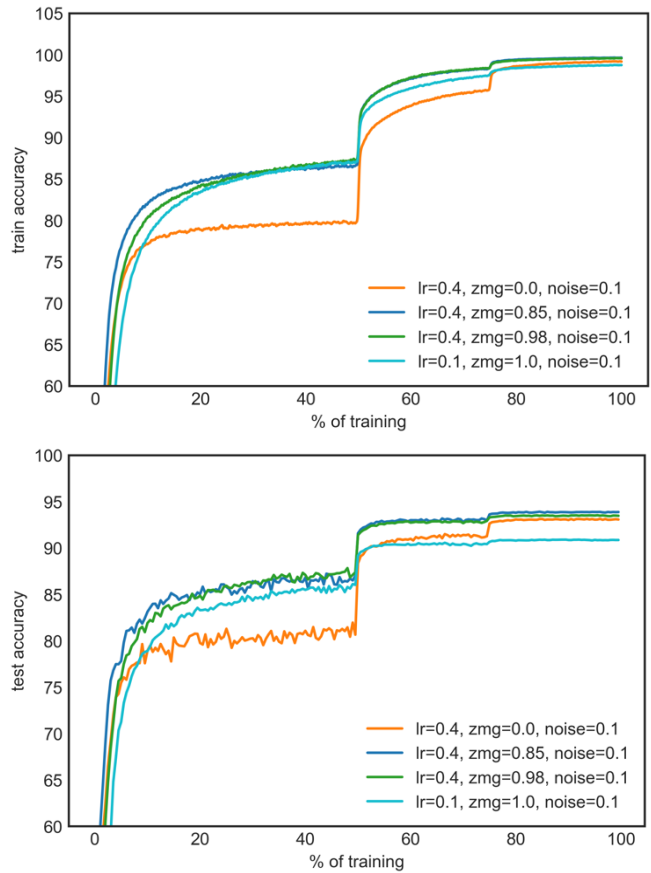

**Figure 3.** The training (top) and test (bottom) accuracies for zmg=0, 0.85, 0.98, and 1 to demonstrate various weighting of the channel-wise gradient mean subtraction

Figure 2 explores the learning rates of 0.1, 0.2, and 0.4 all at the optimal *zmg* factor of 0.85, and without noise injection. The best accuracy is achieved with a learning rate of 0.4 (93.57 ±0.12). This is a little lower than the batch normalization baseline of (93.7 ±0.55). We believe this is lower accuracy is due to the effect of training with minibatch noise which improves generalization. Without such noise, CZMIG cannot quite match the performance of batch normalization.

It can be seen that after each annealing step both the training and test accuracies increase monotonically without any retraction which is in contrast with with batch normalization that has a large rebound. We attribute this to the effective and actual learning rates being the same and being consistent since the filter magnitudes are stabilized near unity by the unity magnitude anchoring loss.

However, with the annealing schedule being tuned for batch normalization baselines, it can be seen that the learning has not hit a plateau yet at the first annealing point. This begs the question of whether delaying the annealing point may have resulted in a higher accuracy, or alternatively whether an even higher learning rate should be applied to increase the rate of learning with the same schedule.

Note that 1D Gaussian smoothing with sigma of 2 is applied to the training accuracy figures only.

Figure 3 explores the channel-wise gradient mean subtraction with *zmg* factors of 0 (i.e. disabled), 0.85, 0.98, and 1 with a fixed learning rate of 0.4 and all with noise injection of amplitude 0.1. The noise simulates the effect of minibatch noise that improves generalization and is explored later. The learning rate is chosen to be comparable to the effective learning rate of the batch normalization baseline. The best accuracy of 94.03 ±0.14 is demonstrated with a *zmg* factor of 0.85. Note that this is 0.33 above the comparable CZMIG baseline without noise injection. The error bar is quite tight showing that the training is highly consistent across trials.

The test accuracy is worse with higher *zmg* of 0.98 dropping to 93.64±0.11, and this rapidly deteriorates with zmg=1 where the model is forced to exactly move along the channel-wise zero mean isocline. At least with this task and architecture, some relaxation of the CZM constraint is needed for optimal training. We do not believe this is true of all architectures which we will explore in future work as spot tests with other architectures and tasks have shown zmg=1 to be as effective as lower values. It may be this is relevant only for weakly supervised tasks such as image classification.

**Figure 4.** The training (top) and test (bottom) accuracies for CZMI+CZMG training with and without noise to demonstrate the effect of noise injection as a regularizer to reduce over-fitting

**Figure 5.** The training (above) and test (below) accuracies of ResNet110 using best CZMIG hyperparameters with noise injection vs the baseline with batch normalization. Standard deviation is indicated by shading around the mean

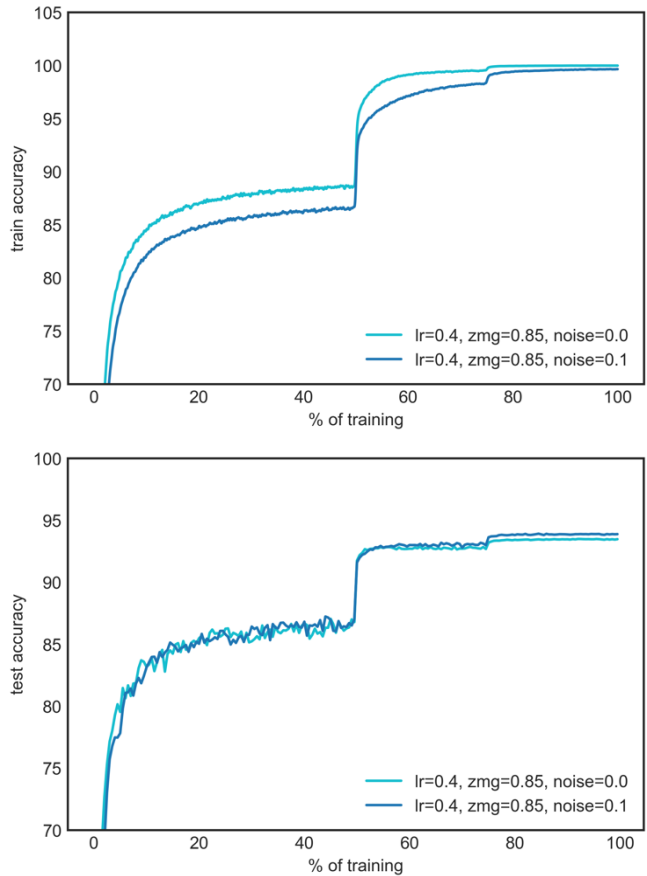

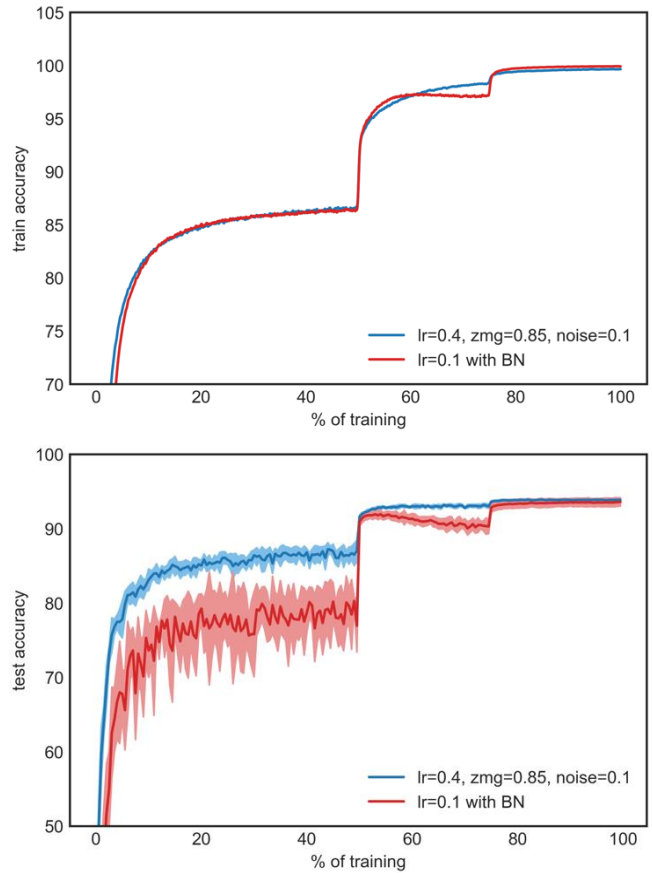

Figure 4 compares the best variant of CZMIG with and without noise injection of 0.1 amplitude. With noise injection there is a clear improvement of 0.46 test accuracy. The improvement is clearer with the training curves. With the test curves the improvement manifests later in training after learning rate annealing. This suggests that with lower learning rate then noise injection is necessary to avoid over-fitting.

Figure 5 (opposite) compares the best CZMIG variant with the batch normalization baseline. The actual learning rate for CZMIG of 0.4 is equivalent to the effective learning rate of 0.4 experienced by the batch normalized training with lr=0.1. The CZMIG result clearly outperforms the batch normalization baseline by a wide margin of 0.3, with a mean accuracy of 94.03 ±0.14 versus 93.70 ±0.55. As with the recent result [26], we demonstrate an improvement over minibatch statistics augmented training. In contrast to [26], we achieve this by training the plain test net without the assistance of normalization. It is worthwhile to note that the variance of the batch normalization experiments is much larger as well.

Also, with batch normalization after the first annealing point the accuracy jump rebounds significantly, while the error range doubles. At the same time the model deflates, and this increases the effective learning rate and explains the effect as this increase in effective learning rate undoes in part the annealing. For CZMIG the magnitude of the filters is tightly controlled throughout training by the L2 unity magnitude anchoring so that effective and actual learning rate are the same.

On a final note, the fact that the net trains beyond state of the art close to the channel-wise zero mean isocline indicates that activations in the DNN are differential in nature as the constant (DC) component is rejected. This may shed light on the internal operation of DNNs as filters respond only to spatially differential signals.

## 6 CONCLUSION

We have demonstrated for the first time that deep convolutional neural nets can be trained stably and to beyond state of the art accuracy at contemporary high learning rates without either normalization or minibatch statistics. We do this by placing the model on the channel-wise zero mean isocline of its filters at the start of training and then during training maintain this condition by adapting the filter gradients by subtracting a fraction of their channel-wise mean which curbs the gradients orthogonal to the isocline. The unmodified test model may then be trained to a significantly higher accuracy versus the batch normalized baseline. The final test accuracy is also far more consistent between experiments which may significantly reduce the experimental burden for research into deep learning generally as there are far fewer outlier results and even a single experiment is representative.

Since the CZM initialization and CZM gradient adaptation are outside of the training graph and do not rely on minibatch statistics or normalization, the memory footprint of the training graph and its computational cost are greatly reduced. For multi-GPU training, no synchronization is required which simplifies training with big data within GPU farms or training with high resolution data.

We further demonstrate that noise injection is vital for improving generalization and we speculate that minibatch noise provides this for batch normalization as a side effect. However, in the case of batch normalization, the amplitude of the noise may not be controlled independently.

Abandoning minibatch statistics should pave the way for future research into recurrent neural nets and online training at high learning rates. We believe that CZMIG is compatible with any network architecture and training regime though we concede the need to explore more architectures and tasks.

We demonstrate that the simplest approach to weight normalization using a unit magnitude uniform distribution works as well as more complex filter initialization methods using fanin and fanout or other factors such as layer position within the net.

The unit magnitude anchoring successfully controls the filter magnitudes during training so that they remain near unity throughout. This avoids the catastrophic deflation experienced with batch normalization that leads to runaway effective learning rate increase in some filters which reduces the effective model capacity. As out method maintains stable filter magnitudes, we conclude that the adaptive effective learning rate that is induced by any normalization-based training technique is not a useful side-effect.

Finally, we note that Filter Response Normalization [26] is based upon the idea that the omission of centering the activations by mean subtraction is mitigated by a thresholded ReLU, TLU. However, in our case no mitigation is needed for this omission as centering is irrelevant in the channel-wise zero mean condition. Further, unreported here we have trained without a bias term without loss of accuracy showing the irrelevance of centering to our method.

## 7 FUTURE WORK

Though we have shown empirically that the channel-wise zero mean isocline is beneficial to stable state of the art training, we lack a rigorous mathematical analysis. We hope to shed more light on this and develop a principled theory and proof.

In developing the method, we have applied it ad hoc to other tasks and network architectures such as semantic segmentation with improved results compared to batch normalization baselines. Very recently, Filter Response Normalization [26] has shown improved accuracy across a range of tasks using activation normalization in conjunction with a trainable thresholded ReLU, and also avoids minibatch statistics. For direct comparison, we are currently actively working to repeat their baselines with our method. In particular, we will add TLU to our method to investigate whether the CZM condition benefits from the adaptive and increased activation range that it may provide versus a ReLU baseline.

## REFERENCES

[1] D. Arpit, V. Campos and Y. Bengio Y. *How to Initialize your Network? Robust Initialization for WeightNorm & ResNets*. arXiv 1906.02341 (2019).

[2] D. Arpit, Y. Zhou, B.U. Kota, B.U. and V. Govindaraju. *Normalization propagation: A parametric technique for removing internal covariate shift in deep networks.* In ICML 2016.

[3] J.L. Ba, J.R. Kiros and G.E. Hinton. *Layer Normalization*. arXiv: 1607.06450 (2016)

[4] J. Bjorck, C. Gomes, B. Selman and K.Q. Weinberger. *Understanding Batch Normalization*. In NeurIPS 2018.

[5] S.R. Bulò, L. Porzi and P. Kontschieder, P. *In-place activated batchnorm for memory-optimized training of DNNa*. arXiv: 1712.02616 (2017).

[6] T. Cooijmans, N. Ballas, C. Laurent, Ç. Gülçehre and A. Courville. *Recurrent batch normalization*. arXiv:1603.09025 (2016).

[7] I. Gitman and B. Ginsburg. *Comparison of batch normalization and weight normalization algorithms for the large-scale image classification.* arXiv: 1709.08145v2 (2017)

[8] X. Glorot and Y. Bengio. *Understanding the difficulty of training deep feedforward neural networks*. In AISTATS, 2010.

[9] B. Hanin and D. Rolnick. *How to start training: The effect of initialization and architecture*. In NeurIPS, 2018.

[10] K. He, Z. Xiangyu, S. Ren and J. Sun. *Delving deep into rectifiers: Surpassing human-level performance on imagenet classification*. In ICCV 2015.

[11] E. Hoffer, R. Banner, R., I. Golan and D. Soudry. *Norm matters: efficient and accurate normalization*. arXiv: 1803.01814v3 (2019).

[12] S. Ioffe and C. Szegedy. *Batch normalization: Accelerating deep network training by reducing internal covariate shift*. ICML 2015.

[13] A. Krogh and J.A. Hertz, *A Simple Weight Decay Can Improve Generalization*. NIPS 1991.

[14] T. van Laarhoven. *L2 Regularization versus Batch and Weight Normalization.* arXiv: 1706.05350 (2017).

[15] B. Poole, S. Lahiri, M Raghu, J. Sohl-Dickstein and S. Ganguli. *Exponential expressivity in deep neural networks through transient chaos*. In NIPS 2016.

[16] M. Quade, M.Abel, N.K. Kutz and S.L. Brunton. *Sparse identification of nonlinear dynamics for rapid model recovery*. arXiv: 1803.0894v2 (2018).

[17] T. Salimans, I. Goodfellow and W. *Zaremba. Improved techniques for training GANs*. NIPS 2016: 2234–2242.

[18] T. Salimans and D.P. Kingma. *Weight normalization: A simple reparameterization to accelerate training of deep neural networks*. NIPS 2016: 901-909.

[19] S. Santurkar, D. Tsipras, A. Ilyas and A. Madry. *How does Batch Normalization help optimization?* In NeurIPS 2018.

[20] A.M. Saxe, J.L. McClelland and S. Ganguli. *Exact solutions to the nonlinear dynamics of learning in deep linear neural networks*. In ICLR 2014.

[21] W. Shang, J. Chiu, J., K. Sohn, K.. *Exploring Normalization in Deep Residual Networks with Concatenated Rectified Linear Units.* AAAI 2017: 1509-1516.

[22] M. Taki. *Deep residual networks and weight initialization.* arXiv:1709.02956 (2017).

[23] G. Yang, J. Pennington, V. Rao, J. Sohl-Dickstein and S. Schoenholz. *A Mean Field Theory of Batch Normalization*. ICLR 2019.

[24] X. Yuan, Z. Feng, M. Norton and X. Li. *Generalized Batch Normalization: Towards accelerating deep neural networks*. In AAAI 2019.

[25] H. Zhang, Y. N. Dauphin and T. Ma. *Fixup Initialization: Residual Learning Without Normalization*. arXiv:1901.09321v2 (2019).

[26] Saurabh Singh and Shankar Krishnan. *Filter Response Normalization Layer: Eliminating Batch Dependence in the Training of Deep Neural Networks*. arXiv:1911.09737v1 (2019).

[27] Siyuan Qiao, Huiyu Wang, Chenxi Liu, Wei Shen, Alan Yuille. *Rethinking Normalization and Elimination Singularity in Neural Networks*. arXiv:1911.09738v1 (2019).

[28] Siyuan Qiao, Huiyu Wang, Chenxi Liu, Wei Shen, Alan Yuille. *Weight Standardization*. arXiv:1903.10520v1 (2019).

[29] Lei Huang, Xianglong Liu, Bo Lang, Dacheng Tao. *Centered Weight Normalization in Accelerating Training of Deep Neural Networks*. ICCV 2017.